\documentclass[openacc]{rstransa}
\usepackage{url}
\usepackage{makecell}

\titlehead{Research}

\begin{document}

\title{Exploring Multi-view Symbolic Regression methods in physical sciences}

\author{
Etienne Russeil$^{1}$, Fabrício Olivetti de França$^{2}$, Guillaume Moinard$^{3}$, Konstantin Malanchev$^{4}$, and Maxime Cherrey$^{5}$}

\address{$^{1}$The Oskar Klein Centre, Department of Astronomy, Stockholm University, AlbaNova,  Stockholm, Sweden\\ 
$^{2}$Universidade Federal do ABC, Santo André, Brazil\\
$^{3}$Sorbonne Université, CNRS, LIP6, Paris, France\\ 
$^{4}$McWilliams Center for Cosmology \& Astrophysics, Department of Physics, Carnegie Mellon University, Pittsburgh, PA 15213, USA\\ 
$^{5}$Independent researcher}

\subject{xxxxx, xxxxx, xxxx}


\corres{Etienne Russeil\\
\email{etienne.russeil@astro.su.se}}

\begin{abstract}
Describing the world behavior through mathematical functions help scientists to achieve a better understand-ing of the inner mechanisms of different phenomena. Traditionally this is done by deriving new equations from first principles and careful observations.
A modern alternative is to automate part of this process with symbolic regression (SR). The SR algorithms search for a function that adequately fits the observed data while trying to enforce sparsity, in the hopes of generating an interpretable equation. A particularly interesting extension to these algorithms is the multi-view symbolic regression (MvSR). It searches for a parametric function capable of describing multiple datasets generated by the same phenomena, which helps to mitigate the common problems of overfitting and data scarcity. Recently, multiple implementations added support to MvSR with small differences between them. In this paper, we test and compare MvSR as supported in Operon, PySR, $\phi$-SO, and eggp, in different real-world datasets. We show that they all often achieve good accuracy while proposing solutions with only few free parameters. However, we find that certain features enable a more frequent generation of better models. We conclude by providing guidelines for future MvSR developments.
\end{abstract}


\begin{fmtext}
\end{fmtext}


\maketitle

\section{Introduction}

Physical sciences have always attempted to describe and predict the world's behavior with as much precision as possible. To that regard, mathematical models, which are able to capture complex phenomena in a simple way often constitutes our highest form of understanding of a given problem. Constrained by their observations, experts have been able to partially explore the space of possible models and construct equations at the center of modern sciences. Over the past decades, new methods have been developed to assist these discoveries.
In particular Symbolic Regression (SR), a type of regression analysis, searches the space of mathematical expressions to find models that best fit a dataset. Unlike traditional regression techniques that assume a predefined model structure, symbolic regression simultaneously discovers both the form and parameters of the model, often yielding interpretable, compact mathematical expressions. This approach is particularly valuable for physical sciences where understanding the underlying physical or causal relationships is as important as predictive accuracy.

A variety of SR techniques have been developed to accomplish the task of exploring the space of possible models. It ranges from the most common approach based on genetic programming \cite{operon, pysr, eggp}, to deep learning \cite{petersen2021deep, PhySO}, random search \cite{McConaghy2011}, or exhaustive search \cite{exhaustive} to cite a few. The versatility of implementations have contributed to the growing adoption of SR techniques across various scientific disciplines. 
Consequently, SR research is no longer reserved to synthetic datasets designed solely for method evaluation, but is increasingly applied to address practical scientific questions using real-world datasets across all fields (e.g. see \cite{sui_cosmo, hernandez2019fast, lacavaFlexibleSymbolicRegression2023a, de2023understanding, cloud_SR}). Nevertheless, despite the growing interest for practical applications, SR have only been confined to single dataset analysis. Indeed the goal is generally to produce a functional form $f(x)$ that minimizes a loss function comparing the predictions to the collected data. Therefore, the solution obtained is fined tuned to the specific dataset it was trained on, and does not necessarily constitute a general law able to describe similar sets of measurements. To address this,  multi-dataset SR analysis have been proposed by multiple authors\cite{multiview, PhySO_Mv, MvSR_epidemic}. In this paper we will refer to these methods as Multi-view Symbolic Regression (MvSR)\footnote{The same concept is also referred as ClassSR in PhySO \cite{PhySO_Mv}, and as Parameterized Template Expressions in PySR \cite{pysr}}. The main idea is that, starting from an ensemble of datasets generated by the same underlying phenomena but observed under a variety of experimental conditions, their goal is to find a unique parametric model $f(x, \theta)$ capable of describing each event independently given the correct optimization of the $\theta$ parameters . The goal of Multi-View Symbolic Regression (MvSR) is to search
for a general parametric model that can simultaneously describe
multiple data sets generated by the same underlying mechanism. In practice, the losses computed independently on each input dataset are aggregated (with $\mathrm{max}$ or $\mathrm{mean}$ for example) into a single value that the SR method tries to minimize. A detailed definition is proposed in \cite{multiview}.

The first studies of this concept show that using a multi-dataset approach leads to improved results in particular for noisy measurements, and favor models with fewer parameters. Beyond numerical advantages, MvSR mitigates the difficulty researchers face in selecting a single representative dataset, particularly in scientific contexts characterized by inherently diverse and variable phenomena. 
This is also related to the idea of factor variables in regression analysis, previously explored with symbolic regression in\cite{factor_mvsr}. 
Currently and to the best of our knowledge, only 4 publicly available SR implementations are able to perform MvSR (Section \ref{sec:imple}). As the concept of multi-view is a broad and general idea, each of them proposes different features specifically related to the multi-dataset nature of the problem. In this study, we propose to evaluate every implementation on a common benchmark (Section \ref{sec:datasets}) composed of real-world use cases from diverse sciences. The goal is not to evaluate their absolute performances, as they will most likely evolve quickly with time, but rather to identify desirable properties and features that should be included in the future. 

\section{Implementations}
\label{sec:imple}

In this section we present each one of the current SR implementations that includes a MvSR approach.  Despite considerable differences in functionality, these implementations share the capability to accept multiple datasets as input and to perform independent parameter optimization on each. We consider these two properties to be the minimum requirement to be considered MvSR.
As described in\cite{multiview}, the loss function of MvSR can be described generically as:

\begin{equation}
    \min_{f}{agg_{i=1 .. k}{\left(\min_{\theta^i}\mathcal{L}(f(x^i; \theta^i), y^i\right)}},
    \label{eq:mvsrfit}
\end{equation}
where the superscript $i$ refers to the index of each dataset, $agg$ is an aggregation function such as $max, avg, med$, and $x \in \mathbb{R}^m$, $\theta \in \mathbb{R}^n$, $n \in \mathbb{N}, l < n < u$, where $m$ is the number of independent variables ($x$), $n$ is the number of parameters in the model ($\theta$) with $n$ being bounded by a finite subset of the natural numbers.\\

\textbf{Operon: }Operon (and its Python's binding PyOperon)\cite{operon} \footnote{\url{https://github.com/heal-research/pyoperon}} is a C\texttt{++} implementation of symbolic regression specifically crafted for high-performance. It is arguably one of the fastest genetic programming implementation for this task, being orders of magnitude faster than the popular approaches \cite{contemp_SR_perf}. It was shown to outperform most SR methods regarding fitting accuracy in various benchmarks~\cite{contemp_SR_perf,de2024srbench++,action}. Operon was extended in~\cite{multiview} to support Multi-view Symbolic Regression allowing to create multiple \emph{evaluators} that fits each dataset independently and aggregate the calculated scores using a choice of aggregation function. It is the only MvSR implementation supporting different aggregation functions such as mean, worst-fitness, best-fitness, median, an harmonic mean. In this analysis, we use the \emph{worst-fitness} function as proposed in the original paper. Unlike the other implementations, Operon does not support the setting of a maximum number of parameters in the expression, and parameters cannot appear more than once per solution. The aggregation function used in this paper for Operon is the \textbf{worst-fitness} value. \\

\textbf{$\phi$-SO: }Physical Symbolic Optimization (hereafter $\phi$-SO \cite{PhySO}) is a Python SR package\footnote{\url{https://github.com/WassimTenachi/PhySO}} focused of physical coherent solutions. As opposed to standard SR approaches that relies on genetic programming, it uses recurrent neural network (RNN) to build equations iteratively generating one token at a time and the next token is \emph{predicted} based on the surrounding information (parent and siblings), units (when available), and additional prior knowledge about the dataset. $\phi$-SO has proven to be highly efficient for exact symbol recovery when all the units are specified. In a follow-up article \cite{PhySO_Mv}, a multiview approach named Class SR has been integrated to $\phi$-SO. Although the name differs, it is equivalent to MvSR. It processes multiple datasets and fits functional forms using realization-specific parameters (parameters - optimized independently for each dataset), and class-parameters (constants - optimized once across all datasets). Within $\phi$-SO, the number of free parameters and constants are customizable hyperparameters, allowing experts to design a solution matching their complexity requirements. We acknowledge that a key advantage of the method resides in its proper management of physical units, but because the datasets proposed in Section \ref{sec:datasets} have no ground truth solution we can not specify units of the parameters. Consequently, $\phi$-SO will be provided with dimensionless values, consistent with the other three implementations. In this implementation the aggregation function used is \textbf{sum}.\\

\textbf{PySR: }PySR \cite{pysr} is a Python and Julia SR package\footnote{\url{https://github.com/MilesCranmer/PySR}} developed specifically for a scientific usage. It builds on traditional genetic programming. It incorporates a multi-population evolutionary algorithm, allowing for independent parallel evolution and therefore enabling efficient multi-thread computation. A recent update of the package - PySR v1.0 - introduced the possibility to input multiple datasets in order to generate a parametric equation \footnote{\url{https://ai.damtp.cam.ac.uk/symbolicregression/stable/examples/template_parametric_expression/\#Parametrized-Template-Expressions}}. It is internally called Parametrized Template Expressions, which corresponds to a MvSR approach. The final loss is computed by summing the losses from all views. PySR enables the choice of the maximum number of parameters to include in the final form as well as the use of constant values that are minimized equally across all datasets. Finally PySR authorizes the same parameter to appear multiple time in the same equation. In this implementation the aggregation function used is \textbf{sum}.\\

\textbf{eggp: }As noted in~\cite{kronberger2024inefficiency}, many GP algorithms suffer from an inefficiency of their search engine as they often revisit and reevaluate the same expressions in their many equivalent forms.
Towards an efficient search, eggp \cite{eggp} \footnote{\url{https://github.com/folivetti/eggp/blob/main/test/mvsr_example.ipynb}} exploits the data structure called e-graph~\cite{willsey2021egg} specifically crafted to compactly store the history of the visited expressions during the search, together with their equivalent forms. For example, if the algorithm generates and evaluates the expression $\log{(x_1 x_2)}$ it will store this expression and $\log{(x_1)} + \log{(x_2)}$ in the visitation history, avoiding unnecessary evaluation and exploration of redundant expressions. The results as reported in~\cite{eggp} showed that this algorithm is more reliable than Operon and PySR on a set of benchmarks. This implementation also supports MvSR allowing the evaluation of multiple datasets and, also admitting the specification of the maximum number of parameters and the reuse of the same parameter in the expression. As in~\cite{multiview}, it uses the \textbf{worst-fitness} fitness as the aggregation.

\section{Datasets}
\label{sec:datasets}

Although many benchmark datasets \cite{contemp_SR_perf}\cite{mastsubara_bench} have been proposed in the SR literature, few are relevant for MvSR approaches, such as in\cite{multiview,PhySO_Mv}. In\cite{PhySO_Mv}, the authors proposed a synthetic benchmark focused on the retrieval of the exact ground-truth expression. In this paper, we follow up with the work in\cite{multiview}, limiting ourselves to real world datasets in order to demonstrate the potential impact of MvSR on concrete problems. This section presents the 5 datasets that have been selected. They originate from various scientific fields and consist of experimental data used by domain experts. Consequently, none of the datasets were generated solely for the purpose of evaluating symbolic regression methods, and no absolute ground-truth solution exists. They cover a range of cases able to challenge the MvSR methods in various ways. We acknowledge that a limitation of the present analysis is its restriction to one-dimensional datasets. This represents an initial step toward the evaluation of MvSR methods, as it facilitates a more intuitive expert interpretation of the results. Future work should extend this assessment to real-world multidimensional datasets.

We also emphasize the absence of data uncertainty in this analysis. Due to the lack of error bars in some of the datasets and the current inability of certain implementations to incorporate uncertainty information, we have chosen to omit uncertainty estimates altogether to maintain consistency across the analysis. However, we are aware of the importance of including uncertainty measurements in symbolic regression exercises \cite{uncertainties} and we stress the need to incorporate proper error bars management in multiview implementations. From the selection of algorithms, only PySR and $\phi$-SO handles uncertainties with a weighted least-squares loss. \\

{\textbf{Supernovae light curves: } Supernovae are extreme stellar explosions that typically marks the end of the life of a star. They are essential to understand as they carry highly valuable information about our universe properties and its history. One method to study them is to construct their light curves -- i.e. their brightness evolution as a function of time -- from which astronomers can extract properties from the supernova.  This extraction typically requires the fit of a phenomenological model to the observations. They are widely used when processing large amounts of light curves, and the question of their optimality is crucial. Since they have been handcrafted by experts, we can reasonably expect that more accurate or simpler models remains to be discovered, enhancing the quality of future analysis. This dataset has the largest samples per view, composed of 6 light curves of 147 points each. It was already used in \cite{multiview} and will be re-evaluated with the recent implementations.

{\textbf{Galaxy rotation curves: }}Dark matter remains one of the most elusive and unresolved puzzles in modern physics. Modern cosmological models that include dark matter have been very successful at describing the large-scale structure of the Universe. However, our current understanding of dark matter fails to fully explain the dynamics of galaxies on smaller scales. Galaxy rotation curves, showing the rotational velocity of matter as a function of radial distance from the galactic center, are a key tool in studying this issue. Therefore, identifying phenomenological models with few parameters is highly valuable \cite{galaxy_salucci}. In this work, we use the co-added rotational curves proposed by \cite{galaxy_dataset} for our multiview analysis. Both the velocity and radial distance axes are normalized. The dataset includes 26 high-quality rotation curves, constructed from over 3200 individual galaxies in the local Universe. It constitutes our sample with the highest number of views, each containing on average 20 data points. \\

{\textbf{Michaelis-Menten dataset: }The Michaelis-Menten dataset \cite{MichaelisMenten} is an historical dataset published in 1913. It has been central to lay the foundation of modern enzyme kinetics. It describes the evolution of the concentration of a final product as enzymes convert sucrose into it. The concentration value is normalized by the initial sucrose concentration such that the evolution starts from 0 and asymptotically reaches 1, meaning that all the sucrose as been converted. This study derived the famous Michaelis-Menten equation, describing the initial velocity of an enzymatic reaction for given conditions. It is composed of 5 views, each generated with a different initial sucrose concentration. Each view has an average of 6 data points, making it the dataset with the least number of observation per view from our ensemble. Therefore, it represents an interesting challenge as overfitting can occur, and a very simple description would be preferred. \\

{\textbf{Nikuradse dataset: }The Nikuradse dataset \cite{Nikuradse} is another well known historical dataset that was produced in 1933. In his study, Nikuradse studies the friction of a liquid in a pipe as a function of the Reynold number of the fluid, i.e. it's level of turbulence. The data offers a with very high sampling and reveals a unique behavior. Over 90 years later, it still remains very difficult to model \cite{unsolved_nikuradse} and has therefore been used multiple times in symbolic regression analysis \cite{nikuradse_Kronberger}\cite{nikuradse_Radwan} as a way to stress test the methods. The dataset can be approached from a multiview perspective, as the experiment has been conducted several times using pipes of increasing roughness. In this context, each pipe is seen as a single view. Therefore, the dataset is composed of 6 views with an average of 60 data points per view. We should notice that the concept of view for this dataset is used only for benchmarking purposes as the roughness of pipes should be interpreted as a continuous value and should ideally be part of the equation. \\

{\textbf{Network degrees distribution: } Networks describe a wide range of phenomena that can be modeled as graphs, where entities or individuals are nodes and interactions among them are links between those nodes. Despite their wide variety, it has been historically stated that their degree distribution is scale-free~\cite{barabasi1999emergence}. In other words, the number of neighbors for each node should follow a power-law distribution. However, more recent studies showed that this behavior is in fact rare~\cite{clauset2009power}. Though empirical distributions are indeed heterogeneous, authors claim an exponential decay describes them as good as, if not better than, the power-law.  In order to contribute to this discussion, we leverage the KONECT\footnote{\url{http://konect.cc/}} project database \cite{konnect}. It provides an impressive number of 1,326 network datasets in 24 different categories that originate from social, biological or technological environments. This open source collection gathers networks that can reach up to 68,349,466 nodes and 2,586,147,869 links for the biggest. We have selected and processed 8 views for our analysis. Each of them has an average of 79 data points. 

\section{Methodology}

In this analysis, we explore the potential of MvSR applied to physical datasets. By evaluating intrinsically different implementations across a range of challenging scenarios, we can not only demonstrate the potential of these methods but also gain insight into their respective strengths and limitations.  We propose to apply every implementation (Section \ref{sec:imple}) on every multiview datasets (Section \ref{sec:datasets}), and evaluate the models generated. Each individual view is randomly partitioned into a training set, comprising 80\% of the data points, and a testing set containing the remaining data points. Model are generated exclusively on the training set. The parameters obtained on the training sample are not re-optimized for the evaluation of the model on the testing sample, hence we measure their capacity to interpolate data points and to predict behaviors outside of the domain range. We use the mean square error (MSE) per degree of freedom as a metric. Hereafter, we call this metric the reduced MSE ($MSE^{*}$) and we define it as,

\begin{equation}
    MSE^{*} = \sum_{i=1}^{n_{\mathrm{obs}}} \frac{(y_i-\hat{y}_i)^2}{(n_{obs}-n_{\mathrm{params}})},
\end{equation}
with $n_{obs}$ being the number of data points, and $n_{params}$ the number of free parameters in the model. We compute it on normalized dataset values $y_{\mathrm{norm}}=y/\mathrm{max}(\mathrm{abs}(y))$, such that the values are comparable across all datasets. This metric not only allow to evaluate the quality of the fit, it also penalizes more complex solutions. For a given dataset, the $MSE^{*}$ is computed independently on every view and aggregated such that the maximum value is used as the final loss. Indeed, the assumption is that all views were generated by a common model, consequently a solution that would fit all views but one is not a general enough model. The $MSE^{*}$ is calculated for the training data and the testing data separately, leading to two metrics. The testing loss is computed without re-minimizing the parameters\footnote{It avoids extra external optimization procedures that could bias results towards functions easier to minimize.} and is therefore a measure of the predictive capacities of MvSR generated models. In this case $n_{params}=0$ and $MSE^{*}=MSE$.

We choose a common set of operators for every dataset [$+$, $-$, $\times$, $\div$, $\mathrm{exp}$, $\mathrm{log}$, $\mathrm{sqrt}$, $\mathrm{abs}$, $\mathrm{pow}$, $\mathrm{square}$]. In addition we evaluate 4 different sets of hyperparameters, varying the maximum size of the equation (i.e. the maximum number of nodes in the expression tree) and the maximum number of free parameters. They are basic hyperparameters, that are common to every MvSR implementations. However, operon does not enable the limitation of parameter number and will therefore run unrestricted. Results will be considered accordingly. Beside this exception, we choose two values of maximum solution size, 15 and 30, and two values for the maximum number of view specific parameters, 2 and 4. Given this, even the largest evaluation, will generate solutions that remains of interpretable size. Indeed, phenomenological models from physical sciences are rarely more complex than 4 parameters, and simplicity often constitutes a key motivation for their practical usage. Hence, we limit our search space to concise solutions. Each set of hyperparameters is evaluated 10 times using different seeds for the random number generation, and different sets of train/test, reducing the role of luck in the analysis. Additionally, we employed the capabilities of PySR and $\phi$-SO to incorporate global constants in order to assess the usefulness of this approach. For $\phi$-SO, the number of constants was limited to two, whereas PySR was allowed an unrestricted number of constants, since the framework does not provide explicit control over this parameter.

It is rigorously impossible to fairly compare the different MvSR implementations. They are fundamentally distinct, employing different algorithms and techniques; consequently, their respective hyperparameter configurations are not directly comparable and may inherently differ. Despite this, we attempt to make the analysis as fair as possible. We chose to leave most hyperparameters to their default values\footnote{\textit{config0b} file is used for default hyperparameters of $\phi$-SO, as recommended for multiview scenarios}, and we only fix the main common hyperparameters. For every method, the number of generation is set to 100, the population size is set to 100 and the number of iterations for the optimization of the parameters is set to 3. Because of limited computing resources, we also set a maximum computation time of 30 min, after which we collect the current best solution. But despite being equal in total number of evaluations, we observe very important gap in computing time (see Table \ref{tab:computation_time}). PySR and $\phi$-SO required an average of $\sim$1000 seconds, while eggp was 20 times faster with an average of $\sim$50 seconds. Finally operon is orders of magnitude faster and required only $\sim$0.1 second per experiment. Therefore, because it can afford much more operations and remain faster than other implementations, and because computation time is a relevant metric for practical usage of MvSR, we include an extra separate setup of operon. We name this setup \textit{operon\_increased}, and we use 1000 generations, a population size of 1000 and 100 optimization iterations. Note that operon is the only method without a constrain on the number of parameters, so it may have an additional advantage to the other methods.

\begin{table}[]
    \centering
    \begin{tabular}{|c|c|c|c|c|c|c}
        \hline
        & PySR & eggp & PhySO & operon & \textit{operon\_increased} \\
        \hline
        Supernovae & 892.9 &	43.3 &	1042.6 & 0.1 & 2.2 \\
        Galaxies &	1785.6 &	83.0 &	801.9 &	0.3 &	7.9 \\
        Network degrees & 574.8 &	45.2 &	601.0 &	0.1 &	2.6 \\
        Nikuradse & 505.2 &	30.7 &	673.0 &	0.1 &	2.2 \\
        Michaelis-Menten &	90.1 &	19.4 &	790.8 &	0.1 &	1.3 \\
        \hline
    \end{tabular}
    \caption{Average computation time of the 10 repetitions of each method and dataset, expressed in seconds.}
    \label{tab:computation_time}
\end{table}

We stress that the exercise proposed here is far from the expected practical usage of MvSR algorithms. Rather than running once with fixed hyperparameters, an expert attempting to build a good functional form will most likely iterate multiple times, fine-tuning the hyperparameters, until a satisfying solution is discovered.  Once an optimal expression is identified, further runs of the MvSR algorithm become unnecessary, thereby justifying the time invested in fine-tuning. In contrast, our experimental scenario is more demanding: hyperparameters were predetermined and remained fixed throughout, with no subsequent optimization. We should consider that, in practice, the user can also try to simplify the final expression either by finding a smaller equivalent form or discarding terms with low impact to the loss.. Therefore, our results reflect a conservative lower bound of what the methods are capable of achieving.

\section{Results}

\subsection{Global metrics}

Figure \ref{fig:general_stat} displays the cumulative distribution of the reduced MSE losses using the \textit{worst-fitness} aggregation function \footnote{using a \textit{mean} aggregation results in the exact same conclusions, see \url{https://github.com/erusseil/mvsr/blob/main/analysis/plots/mean_statistics.pdf}} for each MvSR implementation. The results from all hyperparameter configurations, datasets, and random seeds are stacked on a single histogram, with invalid values (e.g., division by $0$) being treated as infinity. We display reduced MSE values comprised between 0 and $5\times10^{-2}$, corresponding only to the best fit. The distribution is clipped with all values exceeding this threshold aggregated into the $5\times10^{-2}$ bin. The medians of the distribution are displayed in dashed lines. While the exact interpretation of the reduced mean squared error (MSE) is dataset-dependent, a general heuristic of this analysis is to consider any training loss below $5\times10^{-3}$ as indicative of a good model.
\begin{figure}
    \centering
    \includegraphics[width=1\linewidth]{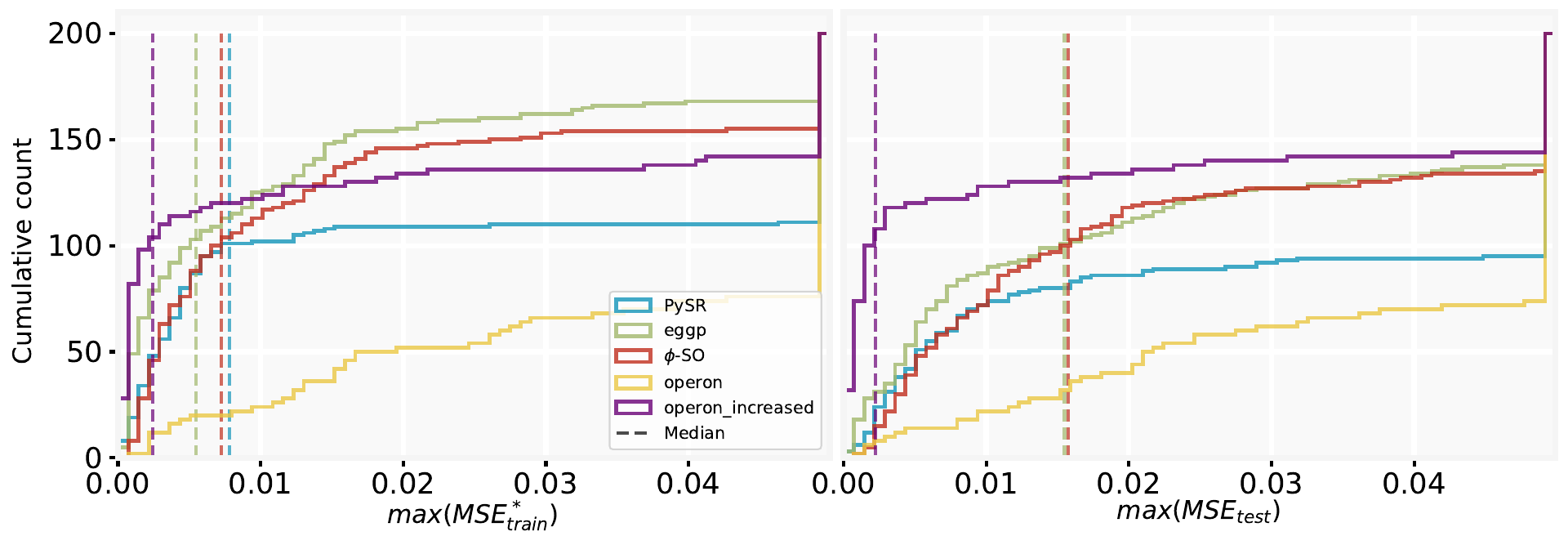}
    \caption{Cumulative distribution of the reduced MSE aggregated using a \emph{worst-fitness} function over all datasets and all hyperparameter configurations. The left panel shows the score on training sample. The right panel shows the score on the testing sample. The distributions are clipped such that every value above is displayed in the $5\times10^{-2}$ bin. The medians of the distributions are showed in dashed lines. Medians not shown are exceeding the clipping threshold.} 
    \label{fig:general_stat}
\end{figure}

To this regard, MvSR methods were able to find high quality models despite the lack of fine tuning. The distributions start with a steep rise driven by the high quality models, followed by a long slow rise. Only operon, when using the same number of evaluations as other methods, is under-performing. This result is expected given the very small computation time used by operon. As demonstrated by \textit{operon\_increased}, this under-performance can easily be overcome by allowing more evaluations (while still largely remaining the fastest method).

The bulk of good answer is not the same for every methods. \textit{operon\_increased} provides well fitting models significantly more often than other implementations. But this result should not be over-interpreted as it is the only method with unrestricted number of parameters, often leading to very over-parametrized solutions. Otherwise we observe that eggp and $\phi$-SO provides similar overall results, with a tendency for eggp to produce slightly more better models. While still able to discover high quality solutions, PySR finds significantly less good models. Except for \textit{operon\_increased}, the results on the testing data show overall significantly worse scores. It does not necessarily indicates an overfit of the models themselves, but rather of the fitted parameters. In that sense, the fits with a significant increase in MSE on the testing set are simply not able to predict unseen data points. However, we notice MvSR is able to generate many models providing a good score on both the training and the testing samples. These models are of high interest for experts.}

Additionally, Figure \ref{fig:pareto_test} displays a trade-off plot between the average number of nodes and the $MSE_{test}$ for all the best solutions generated in the analysis. It summarizes how in most cases, accurate models are discovered by most methods. However the size of the solutions varies significantly from one dataset to another, and no method seems to consistently produce the most accurate and simplest solution. \\

Although looking at global results is useful to understand the frequency of good solutions of the implementations, it is not the most relevant metric. Indeed, an expert using MvSR to discover a model is not interested on the quality of the average answer but purely on the best solution that will be produced. Every other form will be discarded. Therefore we propose to study, for each dataset and for each implementation, the model with the lowest $MSE_{test}$ that was produced. This will provide an overview of the good solution and hint towards the specific features within the implementations that favored the discoveries. Their mathematical expressions are displayed in table \ref{tab:all_equation}. In the following, we will make an individual analysis of the best results for each dataset, showing that there are other criteria to consider when choosing the appropriate model.

\begin{table}[h!]
\centering
\begin{tabular}{|c|c|c|}
\toprule
 & operon & $\phi$-SO \\
\midrule
Supernovae & $\left(\left(e^{A \cdot X}\right)^{B \cdot X}\right)^{0.5}$ & $\left|{\left|{{K_1}^{X} + B}\right|^{A \cdot X}}\right|$\\
\midrule
Galaxies & $A \cdot \left(\left(B + C \cdot e^{D \cdot X}\right)^{4}\right)^{E}$ & $\left|{\left(\frac{A}{\frac{1}{X} \cdot \left(B + \frac{X^{K_1}}{B}\right)}\right)^{C}}\right|$\\
\midrule
\makecell{Michaelis\\Menten} & $A \cdot \left|{X}\right|^{B}$& ${K_1}^{\frac{B^{\frac{\left(B \cdot \left(X + 1\right) + K_1\right)^{0.5}}{A}}}{K_1}}$\\
\midrule
Nikuradse & $\frac{A}{X^{B}} + C \cdot X^{2} + D \cdot X$& $A + e^{X \cdot \left|{\left(A + \log{\left(X \right)} + K_1\right)^{0.5}}\right| - X}$\\
\midrule
\makecell{Network \\ degrees}  & $ A \cdot X - e^{B \cdot X^{2} + C \cdot X} + D$& $\frac{B + K_1}{B + e^{- A^2 + X}}$\\

\toprule
 & PySR & eggp\\
\midrule
Supernovae & $\frac{{K_1}^{\sqrt{{K_2}^{X} + K_3 \cdot \left(K_4 \cdot X + 1\right)^{2}} + \left(B \cdot X + K_5 + \frac{K_6}{A}\right)^{2}}}{K_7}$& $A \cdot \sqrt{\left|{\frac{B}{e^{B^2 \cdot X} + \left|{C + \frac{D}{\left|{e^{B \cdot X}}\right|}}\right|}}\right|}$ \\ 
\midrule
Galaxies &$A + \sqrt{X^{{K_1}^{X} + B}}$ & \makecell{$\left(B \cdot X + D + \frac{C + \sqrt{X} + X \cdot \log{\left(X \right)}}{\sqrt{X}}\right)$ \\$ \times \log{\left(X \right)} + A$}\\
\midrule
\makecell{Michaelis\\Menten} & {\tiny$\left(K_1 \cdot A + \frac{X}{X + e^{\left(X - e^{B}\right)^{A} + \frac{1}{K_2 \cdot B^{K_3}}}} + K_4\right)^{2}$}& $\frac{A \cdot X}{B + \frac{B}{\log{\left(X + \log{\left(\left|{B + \frac{B}{X} + X}\right| \right)} \right)}} + \frac{X \cdot X}{B + \left|{X}\right|}}$\\
\midrule
Nikuradse & $\sqrt{e^{\left|{K_1 + \frac{K_2}{A \cdot e^{X}}}\right|} \cdot \sqrt{\left|{A}\right|} + K_3}$& $\frac{A \cdot X}{\left(A + X^2\right) \cdot \left(\frac{\sqrt{X}}{X} - X\right) - e^{X}} + e^{\frac{B}{B + X}}$\\
\midrule
\makecell{Network \\ degrees}  &{\small \makecell{${K_1}^{e^{B \cdot \left(C + X\right)}} \cdot \left(A + K_2\right)$ }}& $\frac{A + B \cdot X}{C - e^{\sqrt{e^{C + X}}}}$\\
\botrule

\end{tabular}
\caption{Summary table of solutions from the lowest $MSE_{train}$. $K$ represent constant values minimized simultaneously for all views. Other capital letters represents parameters, optimized independently for each view. Their corresponding MSE values are reported in Figs.\ref{fig:supernovae_results} to \ref{fig:graph_results}.}
\label{tab:all_equation}
\end{table}

\begin{figure}
    \centering
    \includegraphics[width=.9\linewidth]{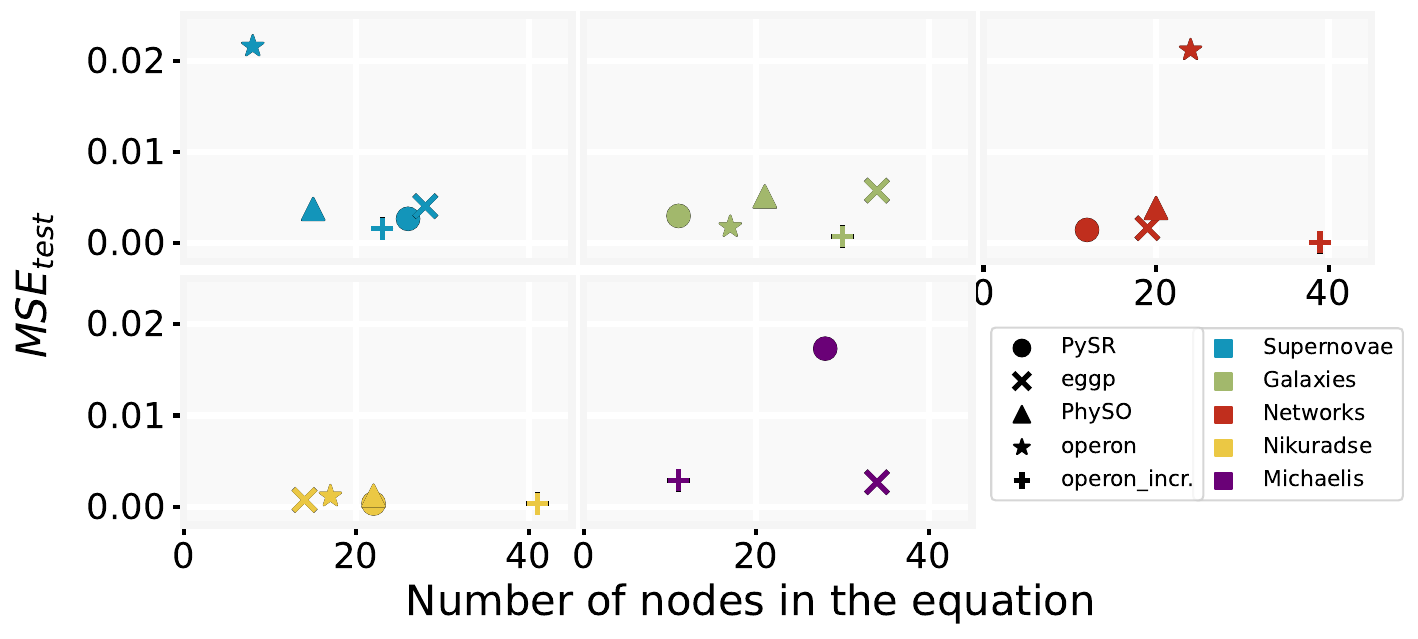}
    \caption{Trade-off plot between the average number of nodes of the most accurate model and the MSE on the test set. The ideal solution should be at the bottom left part of the plot. For readability we display only $MSE_{test}$ values below 0.025. The Michaelis–Menten dataset contains only a limited number of data points; therefore, in this case, the reported $MSE_{test}$ metric is not a reliable indicator of performance.}
    \label{fig:pareto_test}
\end{figure}

\subsection{Best expressions}

\textbf{Supernovae light curves: } Figure \ref{fig:supernovae_results} presents the best models obtained on the supernovae dataset. We observe that two methods did not produce convincing functional forms: operon that generated a symmetrical gaussian-like solution, and eggp that displays an implausible behavior, with the flux exceeding $1$ despite the normalization, which disfavor the solution from a physical point of view. To that regard, PySR and $\phi$-SO provides the best models. They are able to capture the general exponential rise and decay of the light curve and to ensure null flux before and after. In addition, they both only require two parameters, making them competitive with the models commonly used in the field \cite{bazin, Villar_2019}. However, although PySR model is a good fit, it requires 7 constants, resulting from the lack of constraint on the number of constants. While they do not make the model more complex from a minimization point of view, they are very difficult to interpret, and such model would most likely be disregarded by experts. Given that the numerical values of the constants are sometimes similar to each other, it is likely that an equivalent model could be found with less constants. It suggests that even if they are useful for the model precision, there is a need to strictly limit their number. This is illustrated by $\phi$-SO, which provides a simple and interpretable model requiring a single constant value. Finally, we acknowledge that, within the hyperparameter constraints, no models were able to describe the re-brightening observed in some light curves around 30 days after the maximum.\\

\begin{figure}
    \centering
    \includegraphics[width=1\linewidth]{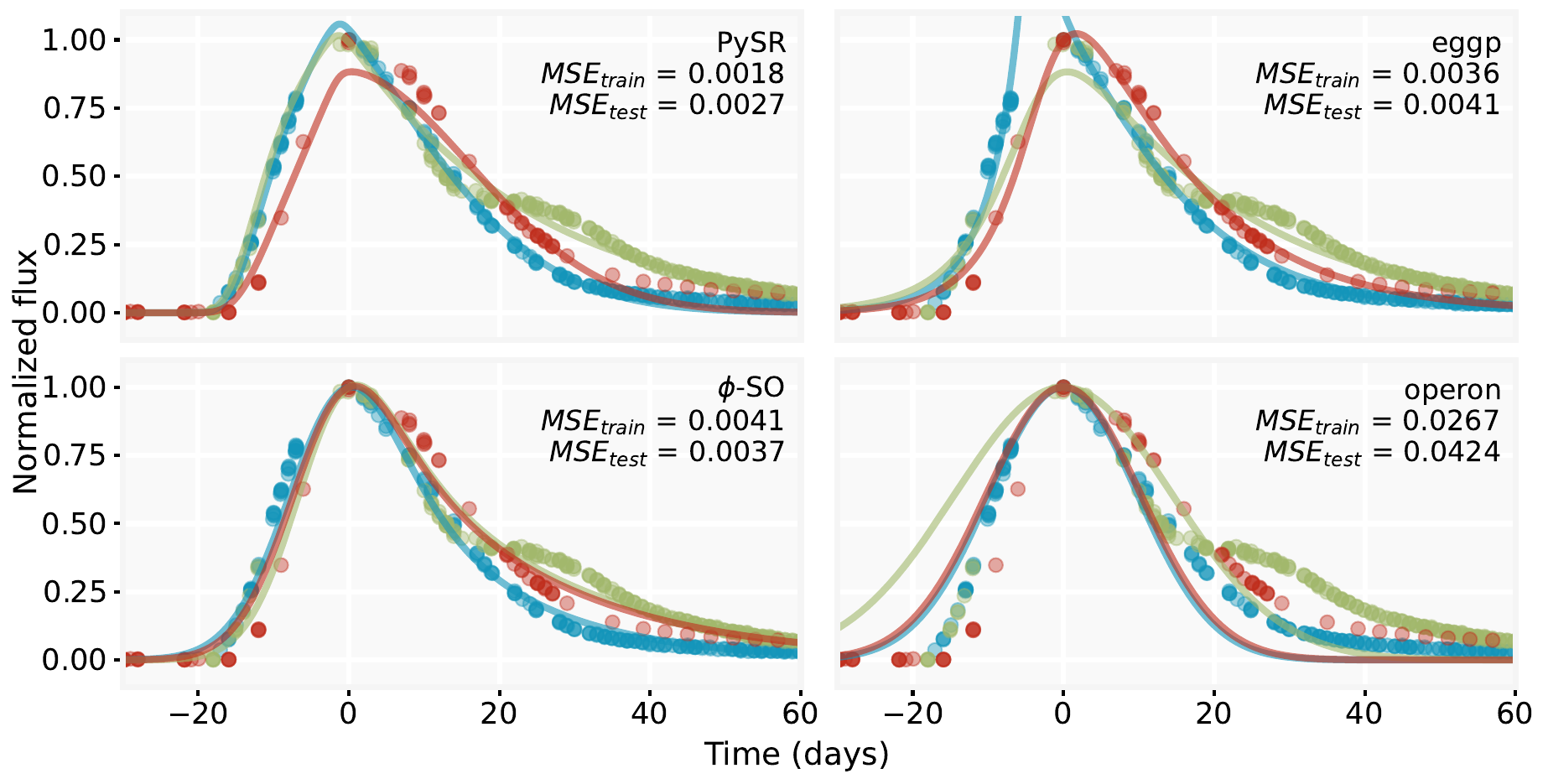}
    \caption{Data fit of the supernovae dataset using the lowest reduced MSE solution from each MvSR implementation. For visual clarity, only 3 views are displayed.}
    \label{fig:supernovae_results}
\end{figure}

\textbf{Galaxy rotation curves: } Figure \ref{fig:galaxy_results} presents the best models obtained on the galaxies dataset. Given that it contains 26 views, the figures display only a minority of the fits, which are not representative of all behaviors. Despite this high number of views, we obtain overall high accuracy from all methods, which are even able to describe galaxies with a slowdown of matter at large distances. The solutions are structurally diverse and range from 2 to 5 parameters. In particular, the model generated by PySR is the most accurate on the test set, while being very economical in terms of number of parameters (2), number of constants (1) and equation size. Finally, we observed a pronounced effect of the large number of views on the computational efficiency of the different methods. More specifically, PySR was substantially slowed down and reached the 30 minutes maximum runtime imposed per experiment.\\

\begin{figure}
    \centering
    \includegraphics[width=1\linewidth]{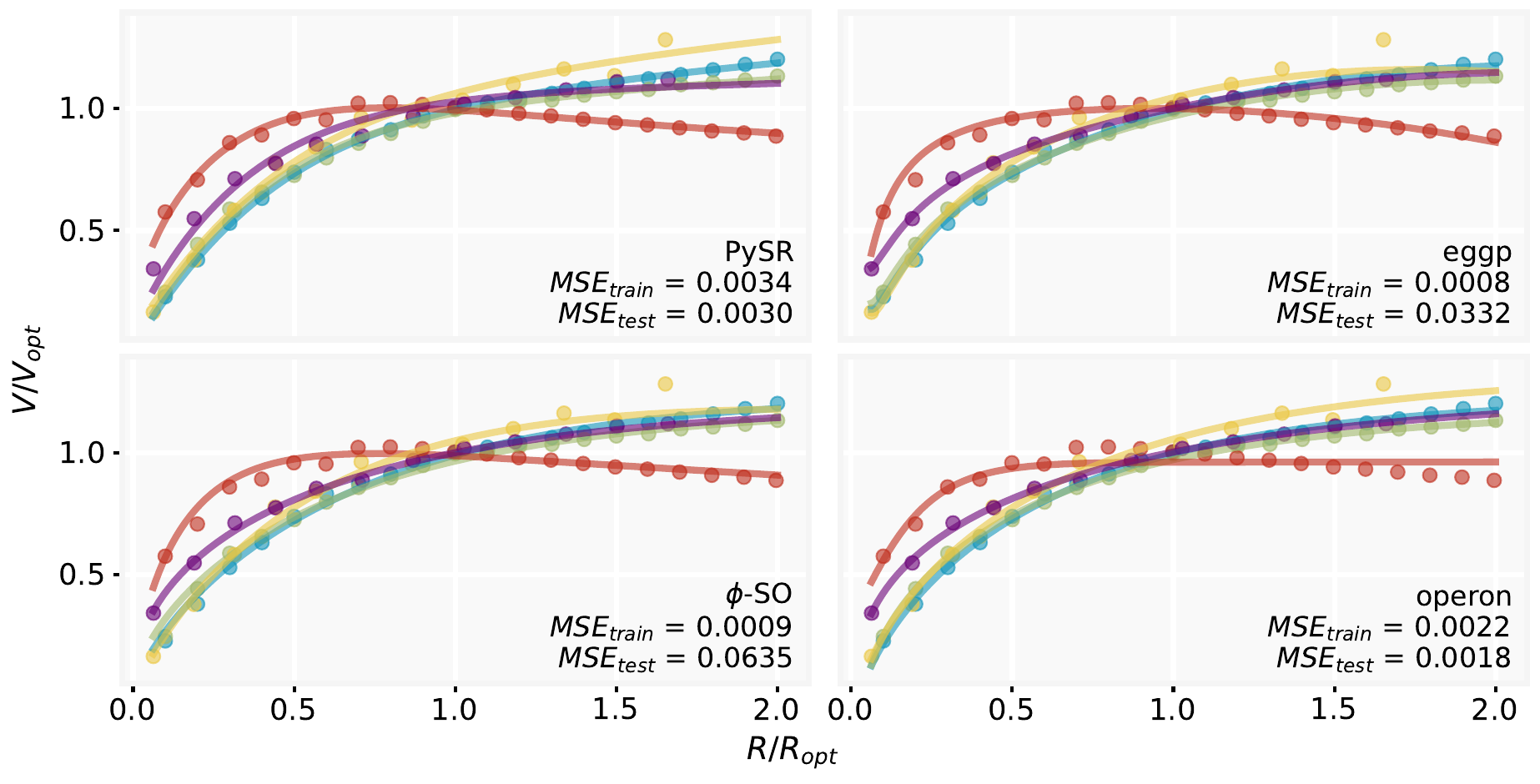}
    \caption{Data fit of the galaxy dataset using the lowest reduced MSE solution from each MvSR implementation. For visual clarity, only 5 views are displayed. Operon's solution includes 5 parameters where other methods were limited to 4.}
    \label{fig:galaxy_results}
\end{figure}

\textbf{Michaelis-Menten dataset: } Figure \ref{fig:bio_results} presents the best models obtained on the Michaelis-Menten dataset. Given the simplicity of the dataset, it is not surprising to see that most methods could build highly accurate models with excellent $MSE_{train}$. Only operon failed to describe the dataset with its limited runtime. Note that given the very low number of measurements, the $MSE_{test}$ is not a reliable metric for this dataset. Despite most models providing good fits, only $\phi$-SO clearly behaves physically plausible given the scientific context. Indeed, the experiment conducted measurements of conversion of sucrose, which intrinsically bound the value between 0 - all sugars are available - and 1 - all sugars have been converted. Hence, $\phi$-SO successfully describes this asymptotic behavior with 2 parameters, and 2 constants. Consequently it is a simple solution, even if its mathematical form is very intricate. The fact that $\phi$-SO constructed an efficient solution using a single constant repeated twice confirms the utility of repeated constants, that would have otherwise required additional complexity.\\

\begin{figure}
    \centering
    \includegraphics[width=1\linewidth]{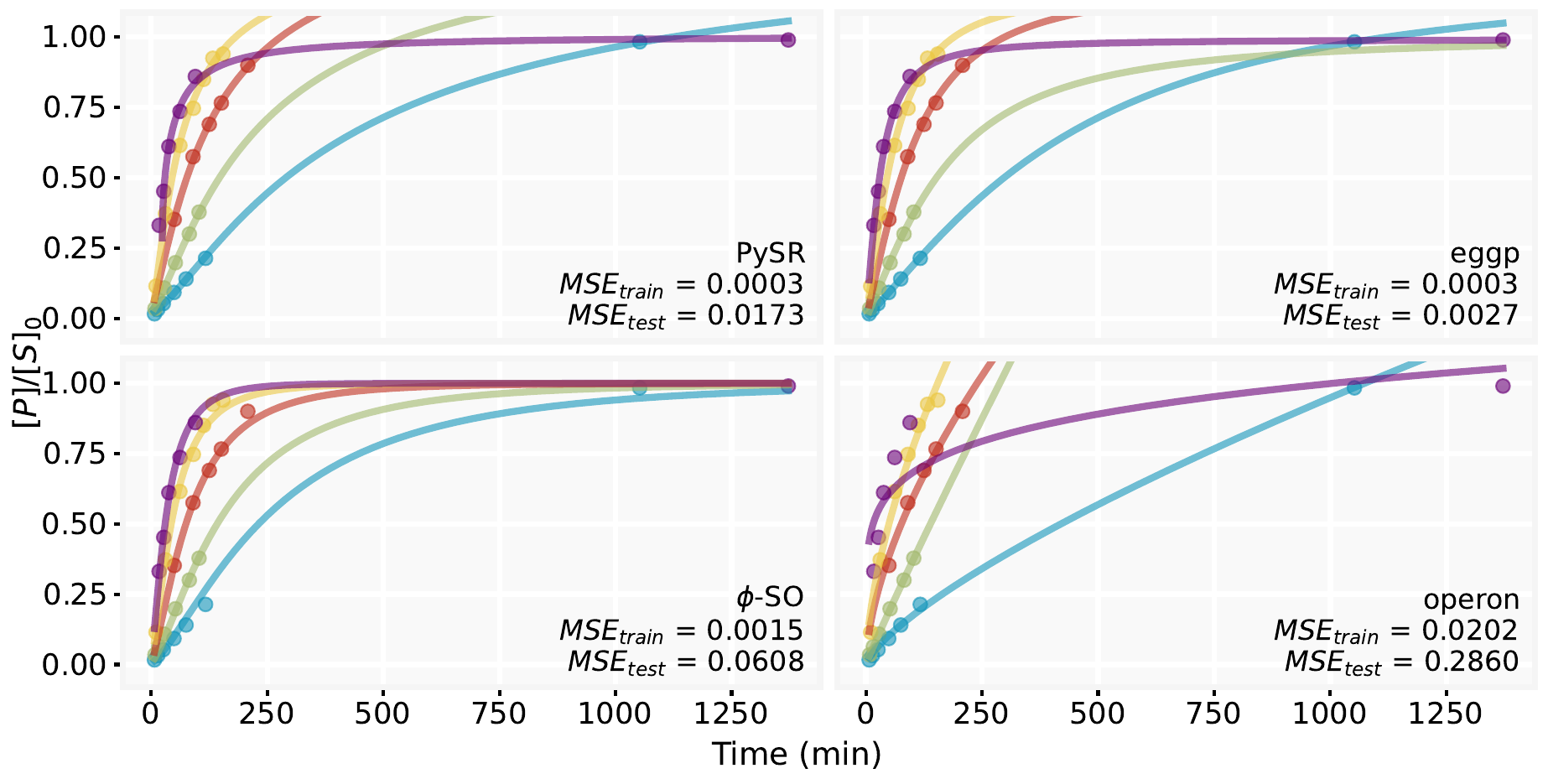}
    \caption{Data fit of the Michaelis-Menten dataset using the lowest $MSE_{train}$ solution from each MvSR implementation.}
    \label{fig:bio_results}
\end{figure}

\textbf{Nikuradse dataset: } Figure \ref{fig:niku_results} presents the best models obtained on the Nikuradse dataset. The $\phi$-SO fit cannot be fully displayed due to the presence of complex-valued outputs. Others methods provide a general accurate fit to the data. Operon and eggp proposes accurate models, but they generally failed to precisely describe the curvature of the views. PySR offers a remarkably simple model. It was able to build an economical model that uses a single parameters and 3 constants, while fitting the data very accurately.\\

\begin{figure}
    \centering
    \includegraphics[width=1\linewidth]{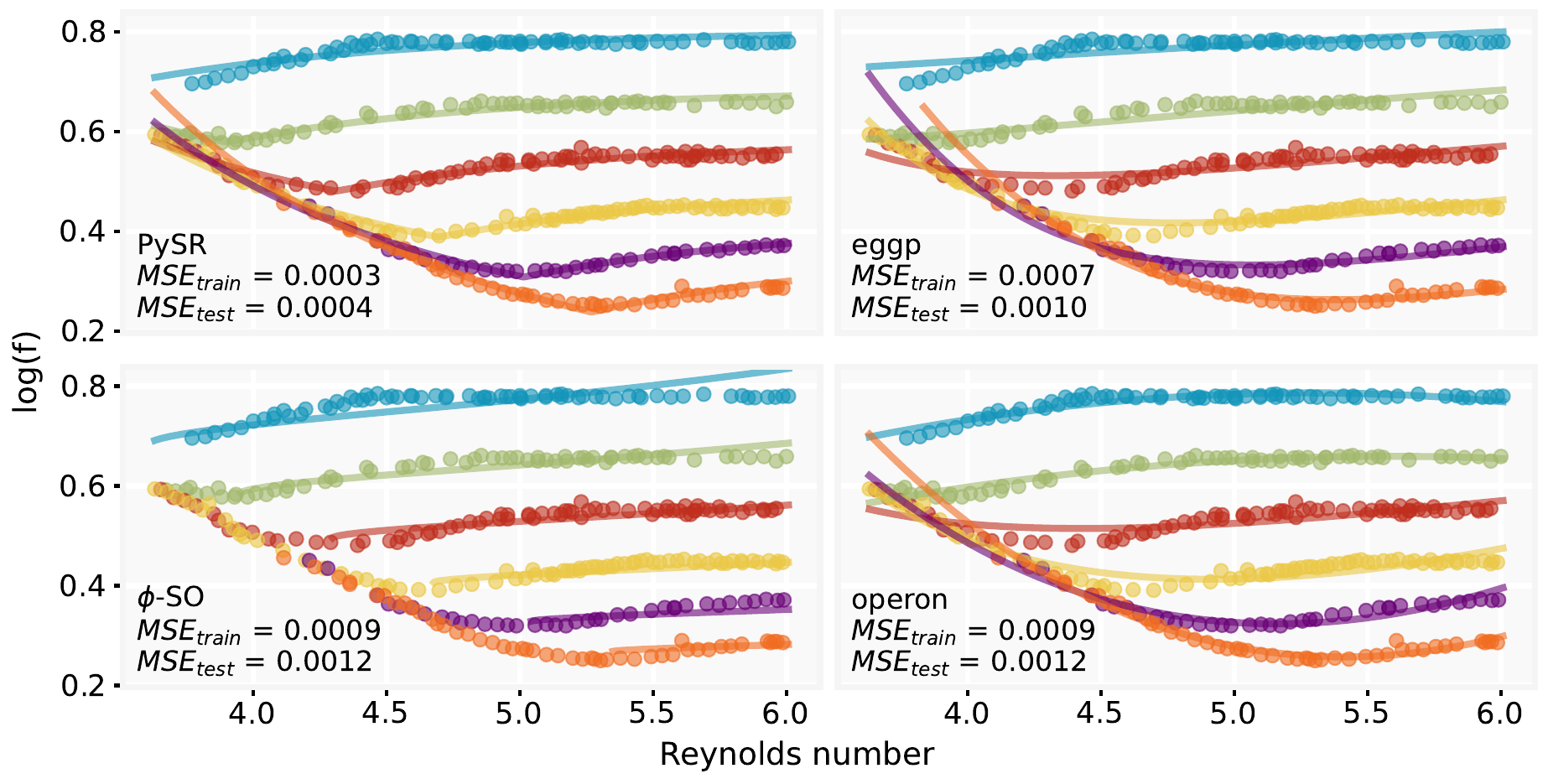}
    \caption{Data fit of the Nikuradse dataset using the lowest reduced MSE solution from each MvSR implementation.}
    \label{fig:niku_results}
\end{figure}

\textbf{Network degrees distribution: } Figure \ref{fig:graph_results} presents the best models obtained on the network distribution dataset. We observe mixed results from the various methods, with a clear failure from operon. $\phi$-SO offers a good fit but is not able model correctly the curvature. However PySR and eggp both produced a very accurate model using 3 parameters (and 2 constants for PySR). The functional forms are relatively simple but include unconventional intricate powers. Besides the first few data points of each view, the solutions are able to describe every part of the behavior of network degrees distributions, making them plausible phenomenological models for this task. \\

\begin{figure}
    \centering
    \includegraphics[width=1\linewidth]{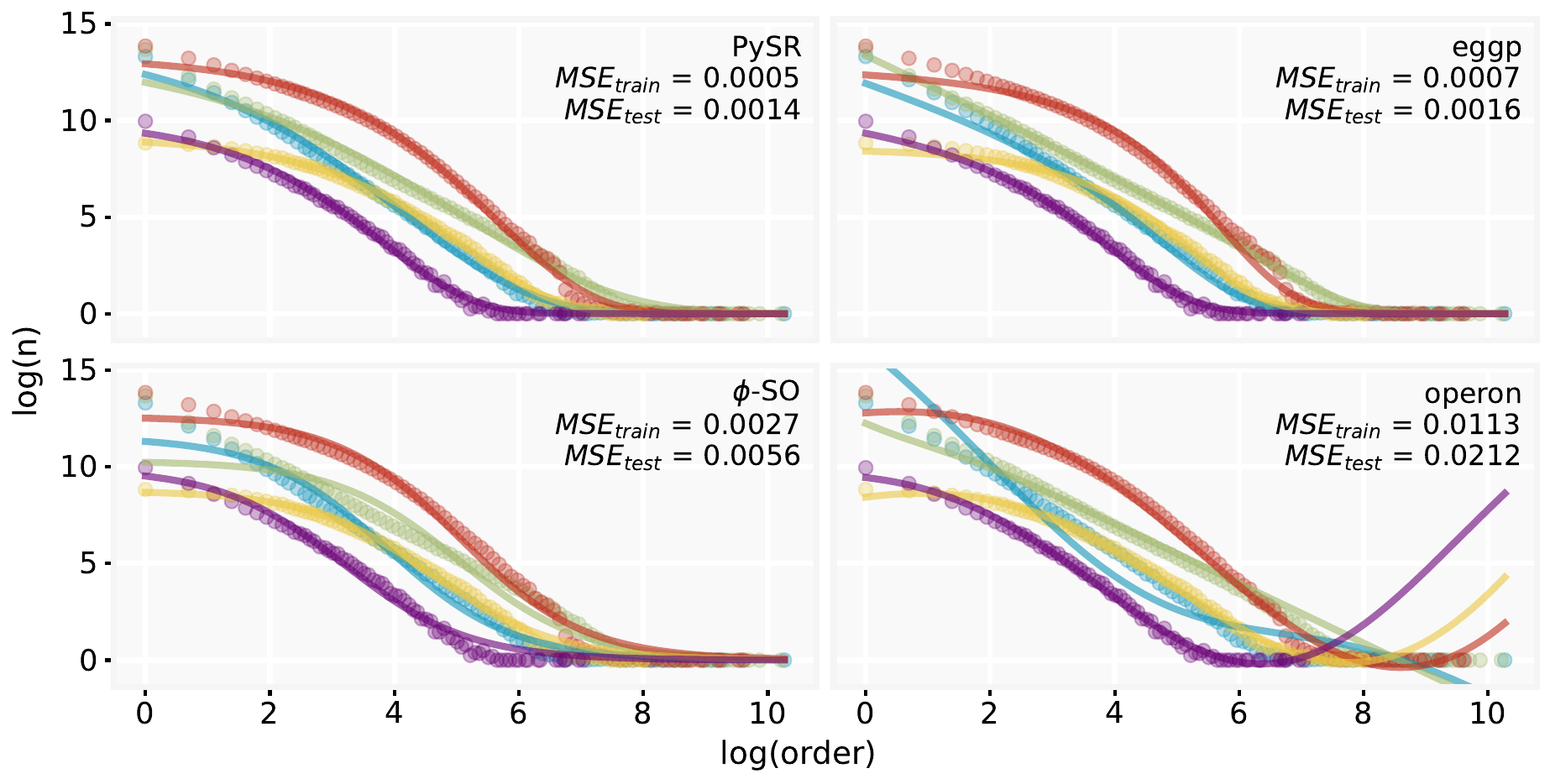}
    \caption{Data fit of the network dataset using the lowest reduced MSE solution from each MvSR implementation. For visual clarity, only 5 views are displayed.}
    \label{fig:graph_results}
\end{figure}

\section{Discussion and conclusion}

The experiments conducted across all datasets, hyperparameter configurations and MvSR implementations have provided some insights concerning this extension of SR. Despite the very constrained experimental conditions, preventing any fine tuning that is often performed for SR techniques, we still obtained a number of plausible models capable of describing challenging real-world datasets. MvSR was able to find functions that are often both accurate and simple. Simplicity constitutes a crucial yet challenging to achieve property for SR, but by combining information from multiple datasets, MvSR reduces overfitting and allocates the parameters to characterize essential properties found among the views as already observed in \cite{multiview,PhySO_Mv}. By examining the successes and failures obtained in our diverse experiments, we have separated several points that are relevant for the improvement of future MvSR implementations. 

\begin{itemize}
    \item \textbf{Parameter control: } Constraining the number of parameters appears to be an essential MvSR feature that drives the search of simpler and relevant models. This is evidence by the ill-behaved and over parameterized models produced by \textit{operon\_increased}, that still achieved the best $MSE^*$ overall. Parameter control can be achieved using a hard constraint, as shown in this study, but ideally MvSR methods should include a loss function that penalizes the degree-of-freedom, such as AIC, BIC, or MDL\cite{exhaustive}.

    \item \textbf{Constant usage: } Many of the most effective models incorporate constants that are jointly optimized across all views. While beneficial in certain cases, our experiment show that the number of constants must also be regulated, either by imposing a maximum or by penalizing their use. PySR, for instance, frequently over-utilizes constants, resulting in models that are harder to comprehend. In contrast, $\phi$-SO illustrates that reducing constant usage can lead to simpler and more interpretable functional forms.  
    \item \textbf{Aggregation strategy: } Our experimental framework neither favored the \emph{worst-fitness} (used by operon and eggp) nor the \emph{mean} (used by PySR and $\phi$-SO) aggregation functions. Further experiment should be conducted to understand whether one is superior to the other. Nevertheless, it appears that this choice should be left to the expert based on the nature of the data used. If all views are equally important, \emph{worst-fitness} should be favored. If, on the contrary, a big sample that may contain unwanted outliers is used, \emph{mean} should be favored. In the future, we advocate that MvSR methods should explicitly set the aggregation function as an hyperparameter.
\end{itemize}

Although multiple experimental configurations were examined in this study, it was not possible to establish a definitive hierarchy among the methods. Indeed, eggp produced the largest number of satisfactory models (per MSE criteria), closely followed by $\phi$-SO, which also yielded several physically coherent models. Although PySR generated fewer satisfactory solutions overall, it produced the highest number of best-performing models across the benchmarks. In terms of efficiency, eggp and operon were the most computationally effective. While operon performed poorly under equal evaluation limits, the \textit{operon\_increased} setting showed that, under more realistic conditions, it can generate the greatest number of accurate models. However they are often over-parameterized due to the current lack of parameter control.

Additionally, the structural variety of models produced shows the relevance of testing the different methods in order to achieve the best result possible. Ideally, each algorithm should allow the exporting of the top-$N$ solutions visited during the search enabling the user to choose the most plausible by other criteria using exploration tools such as \emph{rEGGression}\cite{rEGGression}.}
We also acknowledge the urge for MvSR to properly incorporate uncertainty measurements in their analysis. Currently only PySR and PhySO are able to process this information. Given the relevance of MvSR tools for practical scientific applications, it is crucial that the methods evolves towards more rigorous processing of data uncertainty. Despite the ongoing opportunities for improvement, MvSR methods have already demonstrated significant potential. Given the widespread use of multiple datasets in experimental sciences, MvSR could play a central role in the future development of symbolic regression applied to the physical sciences. All the code and results from the analysis are publicly available at \url{https://github.com/erusseil/mvsr/tree/main}.

\ack{ER is funded by the European Union (ERC, project number 101042299, TransPIre). Views and opinions expressed are however those of the author(s) only and do not necessarily reflect those of the European Union or the European Research Council Executive Agency. Neither the European Union nor the granting authority can be held responsible for them. F.O.F. is supported by Conselho Nacional de Desenvolvimento Cient\'{i}fico e Tecnol\'{o}gico (CNPq) grant 301596/2022-0. Support was provided by Schmidt Sciences, LLC. for K. Malanchev.}


\bibliographystyle{RS}\bibliography{sample}

\begin{thebibliography}{99}

\bibitem{operon}
Burlacu B, Kronberger G, Kommenda M. 2020  Operon C++: An Efficient Genetic Programming Framework for Symbolic Regression. In {\em Proceedings of the 2020 Genetic and Evolutionary Computation Conference Companion} GECCO '20 p. 1562–1570 New York, NY, USA. Association for Computing Machinery.
(\href{http://dx.doi.org/10.1145/3377929.3398099}{10.1145/3377929.3398099})

\bibitem{pysr}
{Cranmer} M. 2023  {Interpretable Machine Learning for Science with PySR and SymbolicRegression.jl}. {\em arXiv e-prints} p. arXiv:2305.01582.
(\href{http://dx.doi.org/10.48550/arXiv.2305.01582}{10.48550/arXiv.2305.01582})

\bibitem{eggp}
de~Franca FO, Kronberger G. 2025  Improving Genetic Programming for Symbolic Regression with Equality Graphs. In {\em Proceedings of the Genetic and Evolutionary Computation Conference} GECCO '25 New York, NY, USA. Association for Computing Machinery.
(\href{http://dx.doi.org/10.1145/3712256.3726383}{10.1145/3712256.3726383})

\bibitem{petersen2021deep}
Petersen BK, Larma ML, Mundhenk TN, Santiago CP, Kim SK, Kim JT. 2021  Deep symbolic regression: Recovering mathematical expressions from data via risk-seeking policy gradients. In {\em International Conference on Learning Representations}.

\bibitem{PhySO}
{Tenachi} W, {Ibata} R, {Diakogiannis} FI. 2023  {Deep Symbolic Regression for Physics Guided by Units Constraints: Toward the Automated Discovery of Physical Laws}. {\em apj} \textbf{959}, 99.
(\href{http://dx.doi.org/10.3847/1538-4357/ad014c}{10.3847/1538-4357/ad014c})

\bibitem{McConaghy2011}
McConaghy T. 2011 pp. 235--260.
In {\em FFX: Fast, Scalable, Deterministic Symbolic Regression Technology}, pp. 235--260. New York, NY: Springer New York.

\bibitem{exhaustive}
{Bartlett} DJ, {Desmond} H, {Ferreira} PG. 2022  {Exhaustive Symbolic Regression}. {\em arXiv e-prints} p. arXiv:2211.11461.
(\href{http://dx.doi.org/10.48550/arXiv.2211.11461}{10.48550/arXiv.2211.11461})

\bibitem{sui_cosmo}
{Sui} C, {Bartlett} DJ, {Pandey} S, {Desmond} H, {Ferreira} PG, {Wandelt} BD. 2024  {syren-new: Precise formulae for the linear and nonlinear matter power spectra with massive neutrinos and dynamical dark energy}. {\em arXiv e-prints} p. arXiv:2410.14623.
(\href{http://dx.doi.org/10.48550/arXiv.2410.14623}{10.48550/arXiv.2410.14623})

\bibitem{hernandez2019fast}
Hernandez A, Balasubramanian A, Yuan F, Mason SAM, Mueller T. 2019  Fast, accurate, and transferable many-body interatomic potentials by symbolic regression. {\em npj Computational Materials} \textbf{5}, 112.
(\href{http://dx.doi.org/10.1038/s41524-019-0249-1}{10.1038/s41524-019-0249-1})

\bibitem{lacavaFlexibleSymbolicRegression2023a}
La~Cava WG, Lee PC, Ajmal I, Ding X, Solanki P, Cohen JB, Moore JH, Herman DS. 2023  A Flexible Symbolic Regression Method for Constructing Interpretable Clinical Prediction Models. {\em npj Digital Medicine} \textbf{6}, 1--14.
(\href{http://dx.doi.org/10.1038/s41746-023-00833-8}{10.1038/s41746-023-00833-8})

\bibitem{de2023understanding}
de~Fran{\c{c}}a FO, di~Genova DVB, Penteado CLC, Kamienski CA. 2023  Understanding conflict origin and dynamics on Twitter: A real-time detection system. {\em Expert Systems with Applications} \textbf{212}, 118748.

\bibitem{cloud_SR}
{Grundner} A, {Beucler} T, {Gentine} P, {Eyring} V. 2024  {Data-Driven Equation Discovery of a Cloud Cover Parameterization}. {\em Journal of Advances in Modeling Earth Systems} \textbf{16}, e2023MS003763.
(\href{http://dx.doi.org/10.1029/2023MS00376310.22541/essoar.168182254.49726852/v1}{10.1029/2023MS00376310.22541/essoar.168182254.49726852/v1})

\bibitem{multiview}
{Russeil} E, {Olivetti de Fran{\c{c}}a} F, {Malanchev} K, {Burlacu} B, {Ishida} EEO, {Leroux} M, {Michelin} C, {Moinard} G, {Gangler} E. 2024  {Multi-View Symbolic Regression}. {\em arXiv e-prints} p. arXiv:2402.04298.
(\href{http://dx.doi.org/10.48550/arXiv.2402.04298}{10.48550/arXiv.2402.04298})

\bibitem{PhySO_Mv}
{Tenachi} W, {Ibata} R, {Fran{\c{c}}ois} TL, {Diakogiannis} FI. 2024  {Class Symbolic Regression: Gotta Fit 'Em All}. {\em apjl} \textbf{969}, L26.
(\href{http://dx.doi.org/10.3847/2041-8213/ad5970}{10.3847/2041-8213/ad5970})

\bibitem{MvSR_epidemic}
{Fajardo-Fontiveros} O, {Mattei} M, {Burgio} G, {Granell} C, {G{\'o}mez} S, {Arenas} A, {Sales-Pardo} M, {Guimer{\`a}} R. 2024  {Machine learning mathematical models for incidence estimation during pandemics}. {\em PLoS Computational Biology} \textbf{20}, e1012687.
(\href{http://dx.doi.org/10.1371/journal.pcbi.1012687}{10.1371/journal.pcbi.1012687})

\bibitem{factor_mvsr}
Kronberger G, Kommenda M, Promberger A, Nickel F. 2018  Predicting friction system performance with symbolic regression and genetic programming with factor variables. In {\em Proceedings of the Genetic and Evolutionary Computation Conference} pp. 1278--1285.

\bibitem{contemp_SR_perf}
{La Cava} W, {Orzechowski} P, {Burlacu} B, {Olivetti de Fran{\c{c}}a} F, {Virgolin} M, {Jin} Y, {Kommenda} M, {Moore} JH. 2021  {Contemporary Symbolic Regression Methods and their Relative Performance}. {\em arXiv e-prints} p. arXiv:2107.14351.
(\href{http://dx.doi.org/10.48550/arXiv.2107.14351}{10.48550/arXiv.2107.14351})

\bibitem{de2024srbench++}
Kommenda M, Majumder M, Cranmer M, Espada G, Ingelse L, Fonseca A, Landajuela M, Petersen B et~al.. 2024  SRBench++: Principled benchmarking of symbolic regression with domain-expert interpretation. {\em IEEE transactions on evolutionary computation}.

\bibitem{action}
Aldeia GSI, Zhang H, Bomarito G, Cranmer M, Fonseca A, Burlacu B, La~Cava WG, de~Fran{\c{c}}a FO. 2025  Call for Action: towards the next generation of symbolic regression benchmark. In {\em Proceedings of the Genetic and Evolutionary Computation Conference Companion} GECCO '25 New York, NY, USA. Association for Computing Machinery.
(\href{http://dx.doi.org/10.1145/3712255.3734309}{10.1145/3712255.3734309})

\bibitem{kronberger2024inefficiency}
Kronberger G, Olivetti~de Franca F, Desmond H, Bartlett DJ, Kammerer L. 2024  The Inefficiency of Genetic Programming for Symbolic Regression. In Affenzeller M, Winkler SM, Kononova AV, Trautmann H, Tu{\v{s}}ar T, Machado P, B{\"a}ck T, editors, {\em Parallel Problem Solving from Nature -- PPSN XVIII} pp. 273--289 Cham. Springer Nature Switzerland.

\bibitem{willsey2021egg}
Willsey M, Nandi C, Wang YR, Flatt O, Tatlock Z, Panchekha P. 2021  Egg: Fast and extensible equality saturation. {\em Proceedings of the ACM on Programming Languages} \textbf{5}, 1--29.

\bibitem{mastsubara_bench}
{Matsubara} Y, {Chiba} N, {Igarashi} R, {Ushiku} Y. 2022  {Rethinking Symbolic Regression Datasets and Benchmarks for Scientific Discovery}. {\em arXiv e-prints} p. arXiv:2206.10540.
(\href{http://dx.doi.org/10.48550/arXiv.2206.10540}{10.48550/arXiv.2206.10540})

\bibitem{uncertainties}
{Olivetti de Franca} F, {Kronberger} G. 2022  {Prediction Intervals and Confidence Regions for Symbolic Regression Models based on Likelihood Profiles}. {\em arXiv e-prints} p. arXiv:2209.06454.
(\href{http://dx.doi.org/10.48550/arXiv.2209.06454}{10.48550/arXiv.2209.06454})

\bibitem{galaxy_salucci}
{Salucci} P, {Persic} M. 1997  {Dark Halos around Galaxies}. In {Persic} M, {Salucci} P, editors, {\em Dark and Visible Matter in Galaxies and Cosmological Implications} vol. 117{\em Astronomical Society of the Pacific Conference Series} p.~1.
(\href{http://dx.doi.org/10.48550/arXiv.astro-ph/9703027}{10.48550/arXiv.astro-ph/9703027})

\bibitem{galaxy_dataset}
{Dehghani} R, {Salucci} P, {Ghaffarnejad} H. 2020  {Navarro-Frenk-White dark matter profile and the dark halos around disk systems}. {\em aap} \textbf{643}, A161.
(\href{http://dx.doi.org/10.1051/0004-6361/201937079}{10.1051/0004-6361/201937079})

\bibitem{MichaelisMenten}
Michaelis L, Menten ML, Johnson KA, Goody RS. 2011  The original Michaelis constant: translation of the 1913 {Michaelis-Menten} paper. {\em Biochemistry} \textbf{50}, 8264--8269.

\bibitem{Nikuradse}
Nikuradse J. 1933  Str{\"o}mungsgestze in rauhen Rohren. Technical Report 361 VDI.
(English translation: NACA Tech. Memo. 1292, National Advisory Commission for Aeronautics, Washington D.C., 1950).

\bibitem{unsolved_nikuradse}
Reichardt I, Pallar\`es J, Sales-Pardo M, Guimer\`a R. 2020  Bayesian Machine Scientist to Compare Data Collapses for the Nikuradse Dataset. {\em Phys. Rev. Lett.} \textbf{124}, 084503.
(\href{http://dx.doi.org/10.1103/PhysRevLett.124.084503}{10.1103/PhysRevLett.124.084503})

\bibitem{nikuradse_Kronberger}
{Kronberger} G, {Olivetti de Franca} F, {Desmond} H, {Bartlett} DJ, {Kammerer} L. 2024  {The Inefficiency of Genetic Programming for Symbolic Regression -- Extended Version}. {\em arXiv e-prints} p. arXiv:2404.17292.
(\href{http://dx.doi.org/10.48550/arXiv.2404.17292}{10.48550/arXiv.2404.17292})

\bibitem{nikuradse_Radwan}
Radwan YA, Kronberger G, Winkler S. 2024  A Comparison of Recent Algorithms for Symbolic Regression to Genetic Programming. .

\bibitem{barabasi1999emergence}
Barab{\'a}si AL, Albert R. 1999  Emergence of scaling in random networks. {\em science} \textbf{286}, 509--512.

\bibitem{clauset2009power}
Clauset A, Shalizi CR, Newman ME. 2009  Power-law distributions in empirical data. {\em SIAM review} \textbf{51}, 661--703.

\bibitem{konnect}
{Kunegis} J. 2014  {Handbook of Network Analysis [KONECT -- the Koblenz Network Collection]}. {\em arXiv e-prints} p. arXiv:1402.5500.
(\href{http://dx.doi.org/10.48550/arXiv.1402.5500}{10.48550/arXiv.1402.5500})

\bibitem{bazin}
{Bazin} G, {Palanque-Delabrouille} N, {Rich} J, {Ruhlmann-Kleider} V, {Aubourg} E, {Le Guillou} L, {Astier} P, {Balland} C, {Basa} S, {Carlberg} RG, {Conley} A, {Fouchez} D, {Guy} J, {Hardin} D, {Hook} IM, {Howell} DA, {Pain} R, {Perrett} K, {Pritchet} CJ, {Regnault} N, {Sullivan} M, {Antilogus} P, {Arsenijevic} V, {Baumont} S, {Fabbro} S, {Le Du} J, {Lidman} C, {Mouchet} M, {Mour{\~a}o} A, {Walker} ES. 2009  {The core-collapse rate from the Supernova Legacy Survey}. {\em aap} \textbf{499}, 653--660.
(\href{http://dx.doi.org/10.1051/0004-6361/200911847}{10.1051/0004-6361/200911847})

\bibitem{Villar_2019}
Villar VA, Berger E, Miller G, Chornock R, Rest A, Jones DO, Drout MR, Foley RJ, Kirshner R, Lunnan R, Magnier E, Milisavljevic D, Sanders N, Scolnic D. 2019  Supernova Photometric Classification Pipelines Trained on Spectroscopically Classified Supernovae from the Pan-STARRS1 Medium-deep Survey. {\em The Astrophysical Journal} \textbf{884}, 83.
(\href{http://dx.doi.org/10.3847/1538-4357/ab418c}{10.3847/1538-4357/ab418c})

\bibitem{rEGGression}
de~Franca FO, Kronberger G. 2025  rEGGression: an Interactive and Agnostic Tool for the Exploration of Symbolic Regression Models. In {\em Proceedings of the Genetic and Evolutionary Computation Conference} GECCO '25 New York, NY, USA. Association for Computing Machinery.
(\href{http://dx.doi.org/10.1145/3712256.3726385}{10.1145/3712256.3726385})

\end{thebibliography}
\end{document}